\documentclass{article}

\usepackage{microtype}
\usepackage{graphicx}
\usepackage{booktabs} % for professional tables

\usepackage{hyperref}

\usepackage[accepted]{icml2024-diffxyz}

\usepackage{listings}
\usepackage[frozencache,cachedir=.]{minted}
\usepackage{amsmath}
\usepackage{amssymb}
\usepackage{mathtools}
\usepackage{amsthm}

\usepackage[capitalize,noabbrev]{cleveref}

\theoremstyle{plain}

\theoremstyle{definition}

\theoremstyle{remark}

\usepackage[textsize=tiny]{todonotes}

\usepackage[
backend=biber,
style=authoryear,  % Added
url=false,
maxcitenames=2
]{biblatex}
\AtBeginBibliography{\small}

\renewbibmacro{in:}{}  % Supress "In:" for @articles 

\usepackage{graphicx}
\usepackage{wrapfig}
\usepackage{subcaption}
\usepackage[font=small]{caption}
\usepackage{booktabs}
\usepackage{multirow}
\usepackage{tabularx}

\usepackage{array}
\newcommand{\PreserveBackslash}[1]{\let\temp=\\#1\let\\=\temp}
\newcolumntype{C}[1]{>{\PreserveBackslash\centering}p{#1}}
\newcolumntype{R}[1]{>{\PreserveBackslash\raggedleft}p{#1}}
\newcolumntype{L}[1]{>{\PreserveBackslash\raggedright}p{#1}}

\usepackage{amsmath}
\usepackage{amssymb}
\usepackage{amsthm}
\usepackage{bm}
\usepackage{dsfont}

\DeclareMathOperator{\E}{\mathbb{E}}

\icmltitlerunning{Revisiting Score Function Estimators for $k$-Subset Sampling}

\begin{document}

\twocolumn[
%\icmltitle{Score Function Estimators for $k$-Subset Sampling}
\icmltitle{Revisiting Score Function Estimators for $k$-Subset Sampling}
%\icmltitle{Are We Overlooking Score Function Estimators for $k$-Subset Sampling?}
%\icmltitle{Score Function Estimators: Another Perspective on Learning with $k$-Subset Sampling}



\icmlsetsymbol{equal}{*}

\begin{icmlauthorlist}
\icmlauthor{Klas Wijk}{kth}
\icmlauthor{Ricardo Vinuesa}{kth}
\icmlauthor{Hossein Azizpour}{kth}
\end{icmlauthorlist}

\icmlaffiliation{kth}{KTH Royal Institute of Technology, Stockholm, Sweden}

\icmlcorrespondingauthor{Klas Wijk}{kwijk@kth.se}

\icmlkeywords{}

\vskip 0.3in
]



\printAffiliationsAndNotice{}  % leave blank if no need to mention equal contribution

\begin{abstract}
    Are score function estimators an underestimated approach to learning with $k$-subset sampling? Sampling $k$-subsets is a fundamental operation in many machine learning tasks that is not amenable to differentiable parametrization, impeding gradient-based optimization. Prior work has focused on relaxed sampling or pathwise gradient estimators. Inspired by the success of score function estimators in variational inference and reinforcement learning, we revisit them within the context of $k$-subset sampling. Specifically, we demonstrate how to efficiently compute the $k$-subset distribution's score function using a discrete Fourier transform, and reduce the estimator's variance with control variates. The resulting estimator provides \emph{both} exact samples and unbiased gradient estimates while also applying to non-differentiable downstream models, unlike existing methods. Experiments in feature selection show results competitive with current methods, despite weaker assumptions.
\end{abstract}

\section{Introduction} \label{sec:introduction}
% Why do we care about subsets?
Subsets are essential in tasks such as feature selection \parencite{balin_concrete_2019, huijben_deep_2019, yamada_feature_2020}, optimal sensor placement \parencite{manohar_data-driven_2018}, learning to explain \parencite{chen_learning_2018}, stochastic $k$-nearest neighbors \parencite{grover_stochastic_2019}, system identification \parencite{brunton_discovering_2016}, and more. Understanding and effectively manipulating subsets is an important step in improving machine methods that model discrete phenomena.

\begin{figure*}
    \centering
    \includegraphics[width=\textwidth]{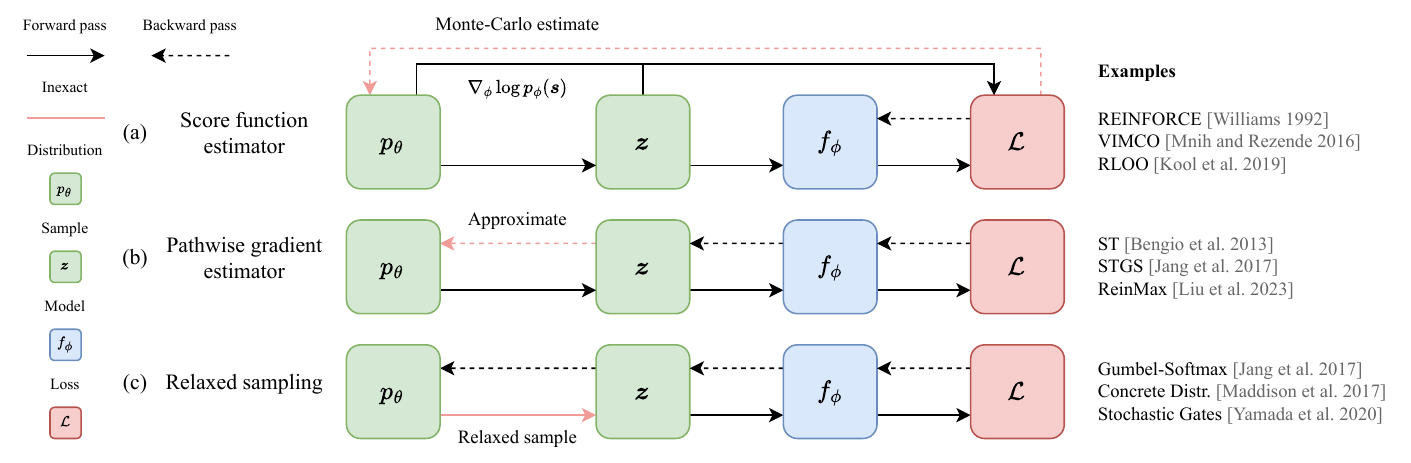}
    \caption{\textbf{Learning by sampling}. Three prominent approaches to learning by sampling: (a) score function estimator, (b) pathwise gradient estimator, and (c) relaxed sampling. We propose a score function estimator for $k$-subset distributions and compare it against existing methods based on approximate pathwise derivatives and relaxed sampling. Because it does not use the pathwise gradient, it is applicable in cases when $f$ is non-differentiable.}
    \label{fig:methods}
\end{figure*}

% Differentiable optimization
A cornerstone of modern machine learning is efficient optimization, typically achieved through differentiable models optimized via stochastic gradient descent. However, not all operations are differentiable, necessitating approximate differentiation to leverage gradient-based optimization. This includes discrete sampling, and thus k-subset sampling which, unlike sampling from Gaussian distributions is not amenable to the reparametrization trick \parencite{kingma_auto-encoding_2014}.

% What's been done for simpler distributions like the categorical distribution?
Differentiable optimization of Bernoulli and categorical distributions have been extensively studied \parencite{bengio_estimating_2013, jang_categorical_2017, maddison_concrete_2017, dimitriev_arms_2021, de_smet_differentiable_2023, liu_bridging_2023}. These distributions are less structured and do not share the combinatorially large support of subset distributions. Still, the methods employed in their optimization serve as a blueprint for more structured distributions.
% What's missing in existing approaches
Existing approaches for differentiable subset sampling \parencite{xie_reparameterizable_2019, ahmed_simple_2023, pervez_scalable_2023} use either relaxed sampling methods or approximate pathwise gradient estimators. While these methods are effective, they produce relaxed samples (which cannot be used in all settings) and biased gradient estimates, respectively. This paper seeks to address these limitations by revisiting score function estimators \parencite{glynn_likelihood_1990, williams_simple_1992, kleijnen_optimization_1996}, a technique well-established in variational inference and reinforcement learning, but overlooked for subset sampling.

% What are we proposing
We propose score function estimators for $k$-subset sampling (SFESS) as a complement to existing methods. This approach is fundamentally different to prior works on $k$-subset sampling, offering both exact samples and unbiased gradient estimates. Furthermore, it does not assume differentiable downstream models, broadening the possible applications of $k$-subset sampling to cases when the downstream model's gradient is unavailable or computationally expensive.

% How did we make it work?
To realize our proposal, we develop an efficient method for computing the score function based on the discrete Fourier transform (DFT) for computing the Poisson binomial distributions' probability density function \parencite{fernandez_closed-form_2010}. Furthermore, we use control variates to significantly reduce the high variance of the vanilla score function estimator.

\section{Method} \label{sec:method}
In this section, we give an overview of our method and provide details on how to compute the score function, reduce the variance with control variates, and anneal the size of the subset to the desired value during training.

%\subsection{Overview}
We are interested in sampling subsets $\bm z$ of size $k$ given a set of $n$ variables. We consider the following conditional distribution:
\begin{align*}
    p_{\bm\theta,k}(\bm z) &= p_{\bm\theta}\left(\bm b \,\big |\, \textstyle \sum_{i=1}^n b_i = k \right) \\&= \frac{\prod_{i=1}^n p_{\bm\theta}(b_i)}{p_{\bm\theta}\left(\textstyle \sum_{i=1}^n b_i = k\right)} \mathds{1}\left[\textstyle\sum_{i=1}^n b_i = k\right],
\end{align*}
where $\bm b \in \{0, 1\}^n$ is independently Bernoulli distributed with parameters $\bm\theta \in [0, 1]^n$ and $\mathds{1}[\cdot]$ denotes the indicator function. This equation induces a particular distribution over the $n \choose k$ possible subsets using only $n$ parameters. Previous work has explored approximate derivatives of this distribution's samples \parencite{xie_reparameterizable_2019, ahmed_simple_2023}. In this work, we instead consider score function estimators that are \emph{exact} in expectation. Hence, we want to compute the score function defined on the region where $\textstyle\sum_{i=1}^n b_i = k$,
\begin{align}
\nabla_{\bm\theta} \log p_{\bm\theta,k}(\bm z) &= \sum_{i=1}^n \nabla_{\bm\theta} \log p_{\bm\theta}(b_i) \label{eq:score-function} \\
&- \nabla_{\bm\theta} \log p_{\bm\theta}\left(\textstyle \sum_{i=1}^n b_i = k\right). \nonumber
\end{align}
Computing the first term is easy, since each $p_{\bm\theta}(b_i)$ is Bernoulli distributed. The second term appears more challenging. Luckily, it follows a Poisson binomial distribution, a generalized binomial distribution where the samples are not necessarily identically distributed. Efficient methods exist for computing the Poisson binomial's density function, including approximate and recursive methods \parencite{le_cam_approximation_1960, wadycki_letters_1973, ahmed_simple_2023}. We follow \textcite{fernandez_closed-form_2010} and compute it using a DFT \parencite{cooley_algorithm_1965} -- leveraging its $\mathcal{O}(n\log n)$ time-complexity and efficient implementation on modern hardware\footnote{We use the Nvidia cuFFT implementation in PyTorch. See \cref{app:score} for pseudocode.}. The gradient of the log probability is computed using automatic differentiation.

Now, being able to compute the score function in \cref{eq:score-function}, we can write the following score function estimator:
\begin{align*}
    &\nabla_{\bm\theta}\E_{p(\bm x)}\E_{p_{\bm\theta,k}(\bm z)} [f_{\bm\phi}(\bm z, \bm x)] \\
    &= \E_{p(\bm x)}\E_{p_{\bm\theta,k}(\bm z)}[\nabla_{\bm\theta}\log p_{\bm\theta,k}(\bm z) f_{\bm\phi}(\bm z,\bm x)] \\
    &\approx \frac{1}{NM}\sum_{i=1}^N\sum_{j=1}^M \nabla_{\bm\theta}\log p_{\bm\theta,k}(\bm z^{(j)}) f_{\bm\phi}(\bm z^{(j)},\bm x^{(i)}), %\label{eq:reinforce-estimator}
\end{align*}
where $N$ samples $\bm x^{(i)} \sim p(\bm x)$ make up the training dataset and $M$ samples $\bm z^{(j)} \sim p_{\bm\theta,k}(\bm z)$ Monte-Carlo estimate the inner expectation. We derive the standard estimator in \cref{app:proof}. 

\paragraph{Efficiently computing the score function} %\label{sec:poibin-pmf}
The second term of \cref{eq:score-function} follows a Poisson binomial distribution. At first glance, the exact computation of this score function is prohibitive due to its combinatorial nature, naïvely expressed as: 
\begin{align*}
    &p_{\bm\theta}\left(\textstyle \sum_{i=1}^n b_i = k\right) = \sum_{\bm b \in \{0, 1\}^n} \mathds{1}\left[\textstyle\sum_i b_i = k\right] p_{\bm\theta}(\bm b).
\end{align*}
\textcite{fernandez_closed-form_2010} derive this closed-form expression using the discrete Fourier transform:

\begin{align}
    &p_{\bm\theta}\left(\textstyle \sum_{i=1}^n b_i = k \right) \label{eq:poibin} \\
    &= \frac{1}{n+1}\operatorname{DFT}\left(\prod_{i=1}^n p_{\bm\theta}(b_i) e^C +\left(1-p_{\bm\theta}(b_i)\right)\right), \nonumber
\end{align}
where $C = 2\sqrt{-1}\pi / (n + 1)$. Note that this expression is solely a function of $\bm\theta$ and $k$ which means we can cache any repeated calls when computing \cref{eq:score-function} with different subsets $\bm z$ with the same size $k$. This is a common occurrence in e.g. instance-wise feature selection \parencite{chen_learning_2018}, where a new $\bm z$ is evaluated for each example $\bm x$.

\paragraph{Reducing variance with control variates}
The vanilla score function estimator generally suffers from high variance. While many variance reduction techniques have been proposed \parencite{mnih_neural_2014, gu_muprop_2016, de_smet_differentiable_2023}, we choose to employ control variates using multiple samples \parencite{mnih_variational_2016, kool_buy_2019} in this work due to its simplicity, unbiasedness, and lack of additional assumptions. The estimator with reduced variance is shown below:
\begin{align*}
    &\nabla_{\bm\theta}\E_{p(\bm x)}\E_{p_{\bm\theta,k}(\bm z)} [f_{\bm\phi}(\bm z, \bm x)] \\
    &\approx \frac{1}{NM}\sum_{i=1}^N\sum_{j=1}^M \nabla_{\bm\theta}\log p_{\bm\theta,k}(\bm z^{(j)}) \\ &\cdot\left( f_{\bm\phi}(\bm z^{(j)},\bm x^{(i)}) - \frac{1}{M - 1}\sum_{k \neq j} f_{\bm\phi}(\bm z^{(k)},\bm x^{(i)})\right).
\end{align*}

\section{Related Work} \label{sec:related-work}
In this section, we provide an overview of current methods for $k$-subset sampling.

\textbf{Relaxed Subset Sampling} \parencite{xie_reparameterizable_2019} \hspace{0.5em} extends the Gumbel-Softmax distribution to distributions over subsets. Despite its elegance, relaxed subset sampling inherits the biased gradient estimation of the Gumbel-Softmax estimator. Furthermore, the top-$k$ sampling procedure sequentially applies the softmax function $k$ times, which limits scalability with respect to $k$ and potentially degrades performance \parencite{pervez_scalable_2023}.

\textbf{SIMPLE} \parencite{ahmed_simple_2023} \hspace{0.5em} approximates the pathwise gradient of the sample using its exact marginals, achieving both lower bias and variance than relaxed subset sampling.

\textbf{Neural Conditional Poisson Subset Sampling (NCPSS)} \parencite{pervez_scalable_2023} \hspace{0.5em} relaxes $k$-subset sampling in a manner different from relaxed subset sampling \parencite{xie_reparameterizable_2019}, allowing subset sizes slightly smaller and larger subsets than $k$. Then, pathwise gradient estimates are used for differentiable optimization. The authors show that NCPSS is more scalable than relaxed subset sampling and that the subset size $k$ can be optimized alongside the distribution's parameters.

\textbf{Implicit Maximum Likelihood Estimation (I-MLE)} \parencite{niepert_implicit_2021} \hspace{0.5em} uses a perturb-and-MAP approach that applies to general optimization problems, with subset sampling as a special case.

% \paragraph{Other methods}
% In some settings, a subset distribution can be modeled using $n$ Bernoulli variables and an auxiliary loss function. An example of this is stochastic gates \parencite{yamada_feature_2020}, a continuous relaxation of the Bernoulli distribution using clipped and mean-shifted Gaussian variables that are differentiable using the chain rule and reparametrization trick \parencite{kingma_auto-encoding_2014}. There are also score function estimators tailored to binary variables, such as Antithetic-REINFORCE-Multi-Sample (ARMS) \parencite{dimitriev_arms_2021}. A similar alternative is to model a subset as the sum of $k$ categorical variables with $n$ categories each. However, this requires $k$ times more parameters than a $k$-subset distribution or independent Bernoulli variables and runs the risk of duplicate inclusions if the categorical distributions are sampled independently \parencite{nilsson_indirectly_2024}. 

\paragraph{Other methods}
In some settings, a subset distribution can be modeled as either the concatenation of $n$ Bernoulli variables or the sum of $k$ categorical variables. This way, a host of gradient estimates for Bernoulli and categorical variables can be used \parencite{yamada_feature_2020, dimitriev_arms_2021, de_smet_differentiable_2023}. However, neither option directly models $k$-subset sampling. Bernoulli variables require some constraint (e.g. a loss term) limiting the subset size \parencite{ahmed_simple_2023}, and categoricals run the risk of duplicate inclusions \parencite{nilsson_indirectly_2024}.
\section{Experiments} \label{sec:experiments}
We consider both feature selection for reconstruction and classification \parencite{balin_concrete_2019, huijben_deep_2019, yamada_feature_2020}. Results on the test set are listed in \cref{tab:results} and the runtimes in \cref{tab:runtime}. Additional plots and details about the experiments are presented in \cref{app:details-of-experiments,app:reconstruction,app:converge}.

\begin{figure*}[ht!]
    \centering
    \includegraphics[width=0.32\textwidth]{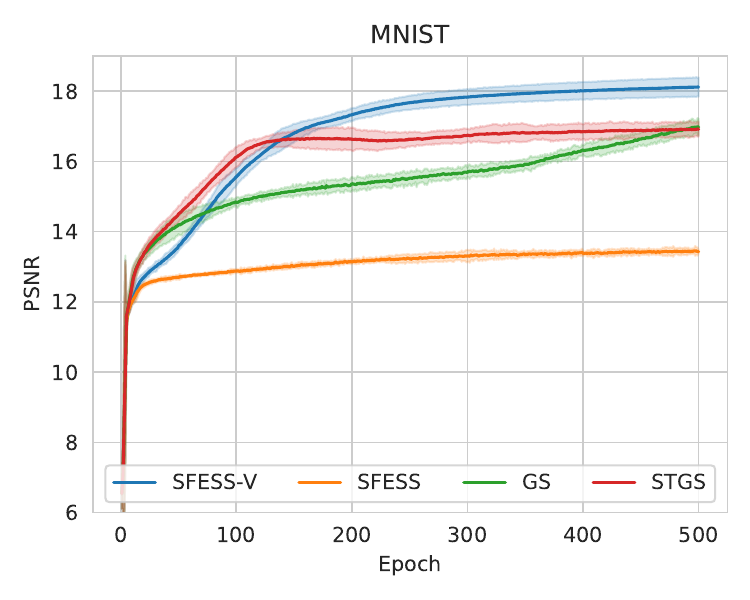}
    \includegraphics[width=0.32\textwidth]{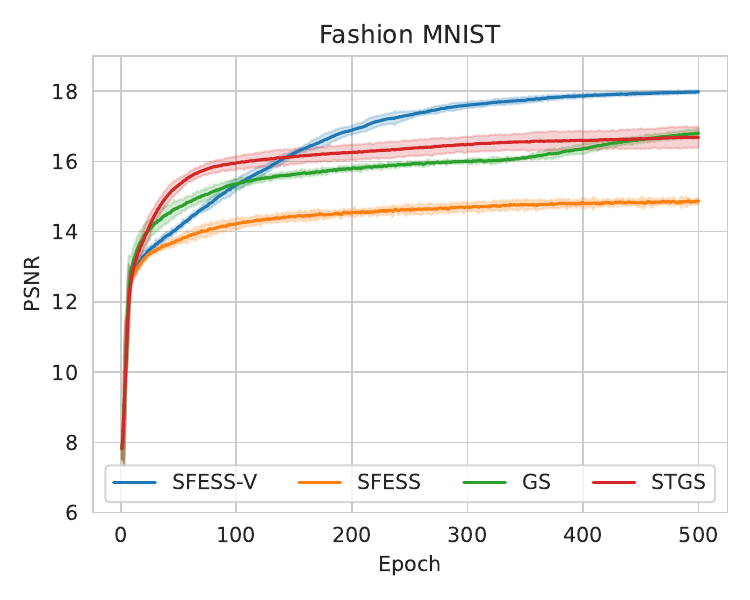}
    \includegraphics[width=0.32\textwidth]{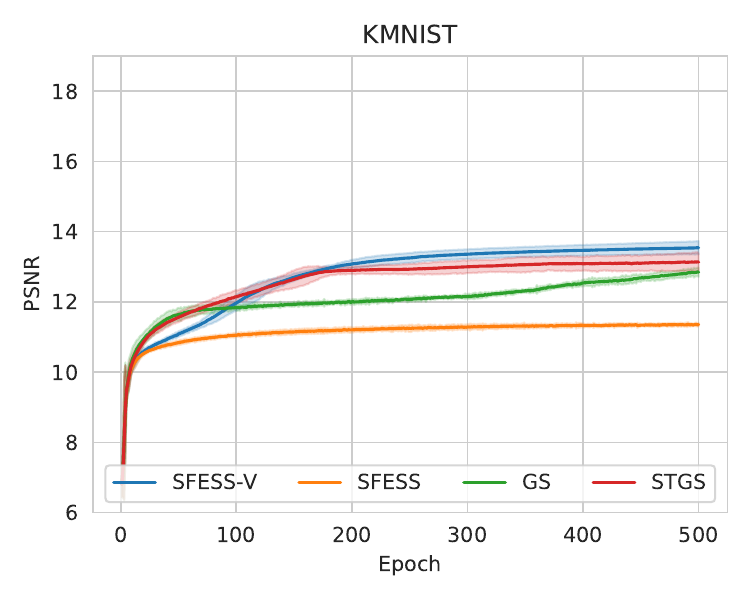}
    \includegraphics[width=0.32\textwidth]{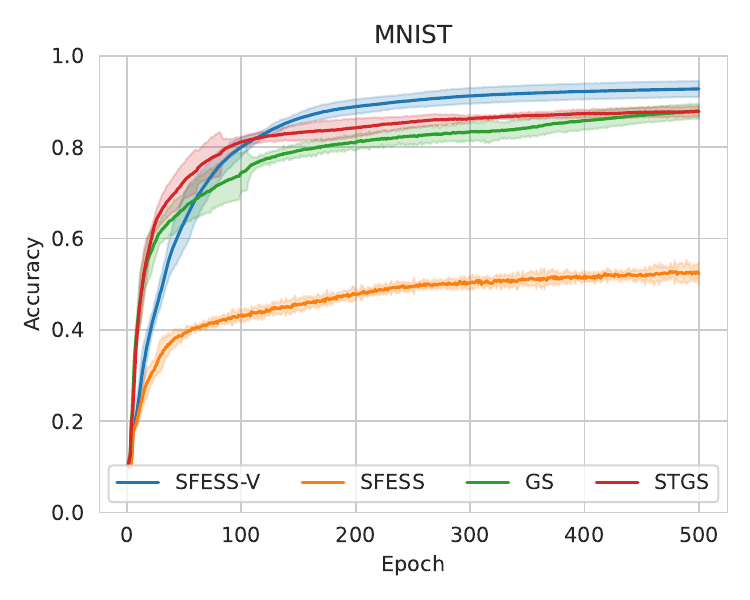}
    \includegraphics[width=0.32\textwidth]{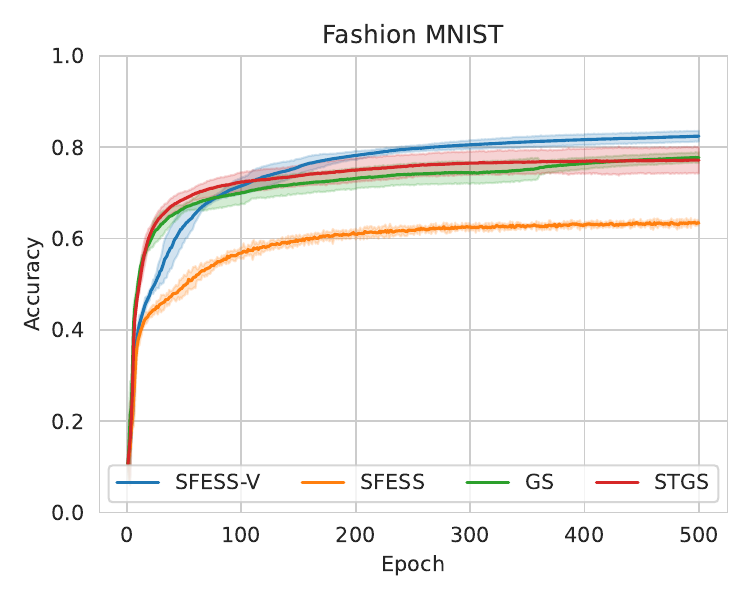}
    \includegraphics[width=0.32\textwidth]{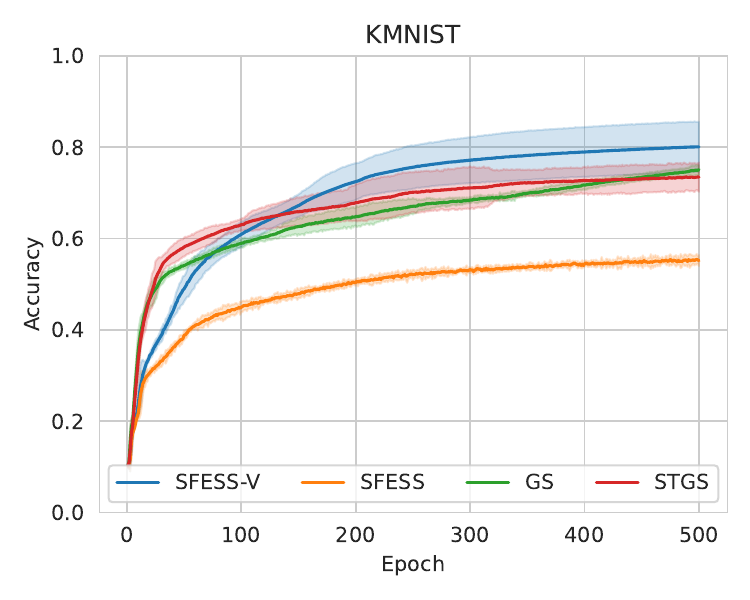}
    \caption{\textbf{Convergence plots}. Convergence of training metrics with $k = 30$ selections. The top row shows reconstruction PSNR and the bottom row classification accuracy. The confidence intervals show two standard deviations computed for 5 repeated runs with different random seeds. See \cref{app:converge} for the corresponding plot with validation data.}
    \label{fig:convergence}
\end{figure*}

\begin{table*}[ht!]
    \centering
    \caption{\textbf{Feature selection results}. Results for feature selection for reconstruction and classification across three datasets with $k = 30$ selections. The confidence intervals show two standard deviations computed for 5 repeated runs with different random seeds. The best mean is shown in \textbf{bold} and the second best mean is \underline{underlined}.}
    \vspace{10pt}
\scalebox{0.8}{
\begin{tabular}{cc C{2.2cm}C{2.2cm}C{2.2cm}C{2.2cm}C{2.2cm}}
\toprule
\multirow{2}{*}{Metric} & 
\multirow{2}{*}{Dataset} & 
\multirow{2}{*}{GS} & 
\multirow{2}{*}{STGS} & 
\multirow{2}{*}{SFESS} &
\multirow{2}{*}{SFESS-V}\\
& & & & & \\
\midrule
                    & MNIST         & \underline{17.402 $\pm$ 0.194} & 17.186 $\pm$ 0.319 & 13.686 $\pm$ 0.412 & \textbf{17.775 $\pm$ 0.209} \\
PSNR $\uparrow$     & Fashion-MNIST & \underline{16.922 $\pm$ 0.289} & 16.642 $\pm$ 0.358 & 15.308 $\pm$ 0.303 & \textbf{17.805 $\pm$ 0.075} \\
                    & KMNIST        & \underline{12.641 $\pm$ 0.083} & 12.561 $\pm$ 0.140 & 11.300 $\pm$ 0.281 & \textbf{12.696 $\pm$ 0.120} \\
\midrule
                    & MNIST         & \underline{0.771 $\pm$ 0.011} & 0.759 $\pm$ 0.319 & 0.416 $\pm$ 0.412 & \textbf{0.796 $\pm$ 0.209} \\
SSIM $\uparrow$     & Fashion-MNIST & \underline{0.586 $\pm$ 0.017} & 0.578 $\pm$ 0.031 & 0.456 $\pm$ 0.016 & \textbf{0.642 $\pm$ 0.011} \\
                    & KMNIST        & 0.428 $\pm$ 0.021 &\textbf{0.464 $\pm$ 0.048} & 0.230 $\pm$ 0.026 & \underline{0.460 $\pm$ 0.022} \\
\midrule
                    & MNIST         & \underline{0.898 $\pm$ 0.014} & 0.898 $\pm$ 0.019 & 0.627 $\pm$ 0.077 & \textbf{0.921 $\pm$ 0.015} \\
Accuracy $\uparrow$ & Fashion-MNIST & \underline{0.777 $\pm$ 0.012} & 0.774 $\pm$ 0.027 & 0.643 $\pm$ 0.112 & \textbf{0.809 $\pm$ 0.009} \\
                    & KMNIST        & \underline{0.604 $\pm$ 0.029} & 0.591 $\pm$ 0.032 & 0.425 $\pm$ 0.023 & \textbf{0.634 $\pm$ 0.059} \\
\bottomrule
\end{tabular}}
    \label{tab:results}
\end{table*}

\begin{table}
    \centering
    \caption{\textbf{Runtime}. Average runtime across all experiments in \cref{tab:results}, all of which have $k = 30$ selections. The measured time includes all surrounding execution, logging, etc. The shortest mean time is shown in \textbf{bold} and the second shortest time is \underline{underlined}.}
    \vspace{10pt}
\scalebox{0.8}{
\begin{tabular}{c C{1.3cm}C{1.3cm}C{1.3cm}C{1.3cm}}
\toprule
\multirow{2}{*}{Metric} & 
\multirow{2}{*}{GS} & 
\multirow{2}{*}{STGS} & 
\multirow{2}{*}{SFESS} &
\multirow{2}{*}{SFESS-V}\\
 & & & & \\
\midrule
Total time & 28.7 min & 29.1 min & \textbf{17.0 min} &  \underline{17.4 min} \\
Time per epoch & 3.45 s & 3.49 s & \textbf{2.04 s} & \underline{2.09 s} \\
\bottomrule
\end{tabular}}
    \label{tab:runtime}
\end{table}

\paragraph{Datasets} 
We evaluate on three datasets: MNIST \parencite{lecun_mnist_1998}, Fashion-MNIST \parencite{xiao_fashion-mnist_2017}, and KMNIST \parencite{clanuwat_deep_2018} split into training, validation, and test sets of sizes 40 000, 10 000, and 10 000.

\paragraph{Baselines} 
We focus our comparison on two baselines from prior work: relaxed sampling using Gumbel-Softmax top-$k$ sampling (GS) and pathwise gradient estimation using the straight-through version of it (STGS) \parencite{xie_reparameterizable_2019}. We compare these baselines against two score function estimator methods: the score function estimator for $k$-subset sampling (SFESS) and the version of the algorithm with control variates (SFESS-V). We use $5$ samples for all methods with SFESS-V using them to to construct control variates, and the other methods reducing variance through averaging.

\paragraph{Discussion}
The experimental results show that SFESS-V is a competitive option for $k$-subset sampling for feature selection. Although GS and STGS converge faster in the early stages of training, SFESS-V gives the best final results, perhaps thanks to its unbiasedness. The weak results of SFESS without control variates demonstrate the detrimental effects of a high-variance gradient estimate and the need for variance reduction. The selections shown in \cref{fig:reconstruction} in \cref{app:reconstruction} further demonstrate this qualitatively.
\section{Conclusion}
In this work, we derived a computationally efficient, unbiased score function estimator for $k$-subset distributions. We then showed how it can be computed using a discrete Fourier transform and how its variance can be reduced using control variates. We emphasize that our proposed estimator is complementary to prior works. An interesting direction for future work is combining score function and pathwise gradient estimators for $k$-subset sampling, as was done for categorical distributions \parencite{tucker_rebar_2017}.

\clearpage
\section*{Acknowledgements}
This work was supported by the Swedish e-Science Research Centre (SeRC). The computations were enabled by the Berzelius resource provided by the Knut and Alice Wallenberg Foundation at the National Supercomputer Centre. Klas Wijk thanks Rickard Maus, Sai bharath chandra Gutha, and Ricky Molén for helpful discussions. We also thank the anonymous reviewers for their valuable feedback and suggestions.
\printbibliography

@inproceedings{pervez_scalable_2023,
	title = {Scalable {Subset} {Sampling} with {Neural} {Conditional} {Poisson} {Networks}},
	booktitle = {International {Conference} on {Learning} {Representations}},
	author = {Pervez, Adeel and Lippe, Phillip and Gavves, Efstratios},
	year = {2023},
}

@inproceedings{ahmed_simple_2023,
	title = {{SIMPLE}: {A} {Gradient} {Estimator} for k-{Subset} {Sampling}},
	booktitle = {International {Conference} on {Learning} {Representations}},
	author = {Ahmed, Kareem and Zeng, Zhe and Niepert, Mathias and Broeck, Guy Van den},
	year = {2023},
}

@inproceedings{de_smet_differentiable_2023,
	title = {Differentiable {Sampling} of {Categorical} {Distributions} {Using} the {CatLog}-{Derivative} {Trick}},
	booktitle = {Advances in {Neural} {Information} {Processing} {Systems}},
	author = {De Smet, Lennert and Sansone, Emanuele and Zuidberg Dos Martires, Pedro},
	year = {2023},
}

@inproceedings{kool_buy_2019,
	title = {Buy 4 {REINFORCE} {Samples}, {Get} a {Baseline} for {Free}!},
	booktitle = {{ICLR} {Workshop} on {Deep} {Reinforcement} {Learning} {Meets} {Structured} {Prediction}},
	author = {Kool, Wouter and Hoof, Herke van and Welling, Max},
	year = {2019},
}

@inproceedings{clanuwat_deep_2018,
	title = {Deep {Learning} for {Classical} {Japanese} {Literature}},
	booktitle = {{NeurIPS} {Workshop} on {Machine} {Learning} for {Creativity} and {Design}},
	author = {Clanuwat, Tarin and Bober-Irizar, Mikel and Kitamoto, Asanobu and Lamb, Alex and Yamamoto, Kazuaki and Ha, David},
	year = {2018},
}

@article{xiao_fashion-mnist_2017,
	title = {Fashion-{MNIST}: a {Novel} {Image} {Dataset} for {Benchmarking} {Machine} {Learning} {Algorithms}},
	journal = {arXiv preprint arXiv:1708.07747},
	author = {Xiao, Han and Rasul, Kashif and Vollgraf, Roland},
	year = {2017},
}

@article{bengio_estimating_2013,
	title = {Estimating or {Propagating} {Gradients} {Through} {Stochastic} {Neurons} for {Conditional} {Computation}},
	journal = {arXiv preprint arXiv:1308.3432},
	author = {Bengio, Yoshua and Léonard, Nicholas and Courville, Aaron},
	year = {2013},
}

@article{brunton_discovering_2016,
	title = {Discovering {Governing} {Equations} from {Data} by {Sparse} {Identification} of {Nonlinear} {Dynamical} {Systems}},
	volume = {113},
	journal = {Proceedings of the National Academy of Sciences},
	author = {Brunton, Steven L. and Proctor, Joshua L. and Kutz, J. Nathan},
	year = {2016},
	pages = {3932--3937},
}

@inproceedings{grover_stochastic_2019,
	title = {Stochastic {Optimization} of {Sorting} {Networks} via {Continuous} {Relaxations}},
	booktitle = {International {Conference} on {Learning} {Representations}},
	author = {Grover, Aditya and Wang, Eric and Zweig, Aaron and Ermon, Stefano},
	year = {2019},
}

@inproceedings{kingma_auto-encoding_2014,
	title = {Auto-{Encoding} {Variational} {Bayes}},
	booktitle = {International {Conference} on {Learning} {Representations}},
	author = {Kingma, Diederik P. and Welling, Max},
	year = {2014},
}

@article{manohar_data-driven_2018,
	title = {Data-{Driven} {Sparse} {Sensor} {Placement} for {Reconstruction}: {Demonstrating} the {Benefits} of {Exploiting} {Known} {Patterns}},
	volume = {38},
	shorttitle = {Data-{Driven} {Sparse} {Sensor} {Placement} for {Reconstruction}},
	journal = {IEEE Control Systems Magazine},
	author = {Manohar, Krithika and Brunton, Bingni W. and Kutz, J. Nathan and Brunton, Steven L.},
	year = {2018},
	pages = {63--86},
}

@article{cooley_algorithm_1965,
	title = {An {Algorithm} for the {Machine} {Calculation} of {Complex} {Fourier} {Series}},
	volume = {19},
	journal = {Mathematics of Computation},
	author = {Cooley, James W. and Tukey, John W.},
	year = {1965},
	pages = {297--301},
}

@article{wadycki_letters_1973,
	title = {Letters to the {Editor}},
	volume = {27},
	journal = {The American Statistician},
	author = {Wadycki, Walter J. and Shah, B. K. and Ghangurde, P. D. and Dudewicz, Edward J. and Mantel, Nathan and Brown, Charles C. and Larson, Harold J. and Barr, Donald R. and Frane, James W. and Saperstein, Bernard and Good, I. J. and Jones, Howard L.},
	year = {1973},
	pages = {123--127},
}

@article{mohamed_monte_2020,
	title = {Monte {Carlo} {Gradient} {Estimation} in {Machine} {Learning}},
	volume = {21},
	journal = {Journal of Machine Learning Research},
	author = {Mohamed, Shakir and Rosca, Mihaela and Figurnov, Michael and Mnih, Andriy},
	year = {2020},
	pages = {1--62},
}

@article{kleijnen_optimization_1996,
	title = {Optimization and {Sensitivity} {Analysis} of {Computer} {Simulation} {Models} by the {Score} {Function} {Method}},
	volume = {88},
	journal = {European Journal of Operational Research},
	author = {Kleijnen, Jack P. C. and Rubinstein, Reuven Y.},
	year = {1996},
	pages = {413--427},
}

@article{williams_simple_1992,
	title = {Simple {Statistical} {Gradient}-{Following} {Algorithms} for {Connectionist} {Reinforcement} {Learning}},
	volume = {8},
	journal = {Machine Learning},
	author = {Williams, Ronald J.},
	year = {1992},
	pages = {229--256},
}

@article{glynn_likelihood_1990,
	title = {Likelihood {Ratio} {Gradient} {Estimation} for {Stochastic} {Systems}},
	volume = {33},
	journal = {Communications of the ACM},
	author = {Glynn, Peter W.},
	year = {1990},
	pages = {75--84},
}

@article{le_cam_approximation_1960,
	title = {An approximation theorem for the {Poisson} binomial distribution},
	volume = {10},
	journal = {Pacific Journal of Mathematics},
	author = {Le Cam, Lucien},
	year = {1960},
	pages = {1181--1197},
}

@article{fernandez_closed-form_2010,
	title = {Closed-{Form} {Expression} for the {Poisson}-{Binomial} {Probability} {Density} {Function}},
	volume = {46},
	journal = {IEEE Transactions on Aerospace and Electronic Systems},
	author = {Fernandez, Manuel and Williams, Stuart},
	year = {2010},
	pages = {803--817},
}

@inproceedings{nilsson_indirectly_2024,
	title = {Indirectly {Parameterized} {Concrete} {Autoencoders}},
	booktitle = {International {Conference} on {Machine} {Learning}},
	author = {Nilsson, Alfred and Wijk, Klas and Gutha, Sai bharath chandra and Englesson, Erik and Hotti, Alexandra and Saccardi, Carlo and Kviman, Oskar and Lagergren, Jens and Vinuesa, Ricardo and Azizpour, Hossein},
	year = {2024},
}

@misc{lecun_mnist_1998,
	title = {The {MNIST} database of handwritten digits},
	url = {http://yann.lecun.com/exdb/mnist/},
	author = {LeCun, Yann and Cortes, Corinna and Burges, Chris},
	year = {1998},
}

@inproceedings{dimitriev_arms_2021,
	title = {{ARMS}: {Antithetic}-{REINFORCE}-{Multi}-{Sample} {Gradient} for {Binary} {Variables}},
	booktitle = {International {Conference} on {Machine} {Learning}},
	author = {Dimitriev, Aleksandar and Zhou, Mingyuan},
	year = {2021},
}

@inproceedings{chen_learning_2018,
	title = {Learning to {Explain}: {An} {Information}-{Theoretic} {Perspective} on {Model} {Interpretation}},
	booktitle = {International {Conference} on {Machine} {Learning}},
	author = {Chen, Jianbo and Song, Le and Wainwright, Martin and Jordan, Michael},
	year = {2018},
}

@inproceedings{balin_concrete_2019,
	title = {Concrete {Autoencoders}: {Differentiable} {Feature} {Selection} and {Reconstruction}},
	booktitle = {International {Conference} on {Machine} {Learning}},
	author = {Balın, Muhammed Fatih and Abid, Abubakar and Zou, James},
	year = {2019},
}

@inproceedings{huijben_deep_2019,
	title = {Deep {Probabilistic} {Subsampling} for {Task}-{Adaptive} {Compressed} {Sensing}},
	booktitle = {International {Conference} on {Learning} {Representations}},
	author = {Huijben, Iris A. M. and Veeling, Bastiaan S. and Sloun, Ruud J. G. van},
	year = {2019},
}

@inproceedings{yamada_feature_2020,
	title = {Feature {Selection} using {Stochastic} {Gates}},
	booktitle = {International {Conference} on {Machine} {Learning}},
	author = {Yamada, Yutaro and Lindenbaum, Ofir and Negahban, Sahand and Kluger, Yuval},
	year = {2020},
}

@inproceedings{liu_bridging_2023,
	title = {Bridging {Discrete} and {Backpropagation}: {Straight}-{Through} and {Beyond}},
	booktitle = {Advances in {Neural} {Information} {Processing} {Systems}},
	author = {Liu, Liyuan and Dong, Chengyu and Liu, Xiaodong and Yu, Bin and Gao, Jianfeng},
	year = {2023},
}

@inproceedings{niepert_implicit_2021,
	title = {Implicit {MLE}: {Backpropagating} {Through} {Discrete} {Exponential} {Family} {Distributions}},
	booktitle = {Advances in {Neural} {Information} {Processing} {Systems}},
	author = {Niepert, Mathias and Minervini, Pasquale and Franceschi, Luca},
	year = {2021},
}

@inproceedings{mnih_neural_2014,
	title = {Neural {Variational} {Inference} and {Learning} in {Belief} {Networks}},
	booktitle = {International {Conference} on {Machine} {Learning}},
	author = {Mnih, Andriy and Gregor, Karol},
	year = {2014},
}

@inproceedings{gu_muprop_2016,
	title = {{MuProp}: {Unbiased} {Backpropagation} for {Stochastic} {Neural} {Networks}},
	booktitle = {International {Conference} on {Learning} {Representations}},
	author = {Gu, Shixiang and Levine, Sergey and Sutskever, Ilya and Mnih, Andriy},
	year = {2016},
}

@inproceedings{mnih_variational_2016,
	title = {Variational inference for {Monte} {Carlo} objectives},
	booktitle = {International {Conference} on {Machine} {Learning}},
	author = {Mnih, Andriy and Rezende, Danilo J.},
	year = {2016},
}

@inproceedings{jang_categorical_2017,
	title = {Categorical {Reparameterization} with {Gumbel}-{Softmax}},
	booktitle = {International {Conference} on {Learning} {Representations}},
	author = {Jang, Eric and Gu, Shixiang and Poole, Ben},
	year = {2017},
}

@inproceedings{maddison_concrete_2017,
	title = {The {Concrete} {Distribution}: {A} {Continuous} {Relaxation} of {Discrete} {Random} {Variables}},
	booktitle = {International {Conference} on {Learning} {Representations}},
	author = {Maddison, Chris J. and Mnih, Andriy and Teh, Yee Whye},
	year = {2017},
}

@inproceedings{tucker_rebar_2017,
	title = {{REBAR}: {Low}-variance, unbiased gradient estimates for discrete latent variable models},
	booktitle = {Neural {Information} {Processing} {Systems}},
	author = {Tucker, George and Mnih, Andriy and Maddison, Chris J. and Lawson, Dieterich and Sohl-Dickstein, Jascha},
	year = {2017},
}

@inproceedings{xie_reparameterizable_2019,
	title = {Reparameterizable {Subset} {Sampling} via {Continuous} {Relaxations}},
	booktitle = {International {Joint} {Conference} on {Artificial} {Intelligence}},
	author = {Xie, Sang Michael and Ermon, Stefano},
	year = {2019},
}

%%%%%%%%%%%%%%%%%%%%%%%%%%%%%%%%%%%%%%%%%%%%%%%%%%%%%%%%%%%%%%%%%%%%%%%%%%%%%%%
%%%%%%%%%%%%%%%%%%%%%%%%%%%%%%%%%%%%%%%%%%%%%%%%%%%%%%%%%%%%%%%%%%%%%%%%%%%%%%%
% APPENDIX
%%%%%%%%%%%%%%%%%%%%%%%%%%%%%%%%%%%%%%%%%%%%%%%%%%%%%%%%%%%%%%%%%%%%%%%%%%%%%%%
%%%%%%%%%%%%%%%%%%%%%%%%%%%%%%%%%%%%%%%%%%%%%%%%%%%%%%%%%%%%%%%%%%%%%%%%%%%%%%%
\clearpage
\appendix
\onecolumn

\section{Details of Experiments} \label{app:details-of-experiments}
This appendix gives additional information on network architecture, initialization, hyperparameters and the computing environment used in the experiments.

\subsection{Network Architecture}
The network architectures used in the main feature selection experiments for reconstruction and classification are shown in \cref{tab:reconstruction-network,tab:classification-network} respectively. We used a convolutional neural network for reconstruction and a fully connected network with dropout for classification.

\begin{table}[ht!]
    \centering
    \caption{\textbf{Reconstruction network}. All convolutions have stride 1.}
    \vspace{10pt}
    \scalebox{0.85}{
    \begin{tabular}{cccc}
    \toprule 
    \multirow{2}{*}{Type} & 
    \multirow{2}{*}{\#Features in} & 
    \multirow{2}{*}{\#Features out} &
    \multirow{2}{*}{Activation} \\
    & & \\
    \midrule
    Reshape             & $28 \times 28 \times 1$  & $748$                    & --      \\
    Linear              & $748$                    & $748$                    & ReLU    \\
    Linear              & $748$                    & $748$                    & --      \\
    Reshape             & $748$                    & $28 \times 28 \times 1$  & --      \\
    Conv ($3 \times 3$) & $28 \times 28 \times 1$  & $28 \times 28 \times 16$ & ReLU    \\
    Conv ($3 \times 3$) & $28 \times 28 \times 16$ & $28 \times 28 \times 16$ & ReLU    \\
    Conv ($3 \times 3$) & $28 \times 28 \times 16$ & $28 \times 28 \times 1$  & Sigmoid \\
    \bottomrule
    \end{tabular}}
    \label{tab:reconstruction-network}
\end{table}

\begin{table}[ht!]
    \centering
    \caption{\textbf{Classification network}}
    \vspace{10pt}
    \scalebox{0.85}{
    \begin{tabular}{cccc}
    \toprule 
    \multirow{2}{*}{Type} & 
    \multirow{2}{*}{\#Features in} & 
    \multirow{2}{*}{\#Features out} &
    \multirow{2}{*}{Activation} \\
    & & \\
    \midrule
    Reshape             & $28 \times 28 \times 1$  & $748$                    & --      \\
    Linear              & $748$                    & $256$                    & ReLU    \\
    Dropout ($20\%$)    & --                       & --                       & --      \\
    Linear              & $256$                    & $256$                    & ReLU    \\
    Dropout ($20\%$)    & --                       & --                       & --      \\
    Linear              & $256$                    & $256$                    & ReLU    \\
    Dropout ($20\%$)    & --                       & --                       & --      \\
    Linear              & $256$                    & $10$                     & Softmax \\
    \bottomrule
    \end{tabular}}
    \label{tab:classification-network}
\end{table}

\subsection{Initialization}
The subset distribution parameters were uniformly initialized to $k / n$. For the network parameters, we used the default initialization in PyTorch.

\subsection{Optimization and Hyperparameters}
We use the Adam optimizer with the momentum parameters $\alpha_{\bm\theta} = (0.99, 0.999)$ for the selector and $\alpha_{\bm\phi} = (0.9, 0.999)$ with weight decay $10^{-4}$ for the downstream network. We set the batch size to $1024$. Learning rates were set to $10^{-2}$ for the selector and $10^{-4}$ for the downstream network and all models are trained for $500$ epochs. We use cross-entropy loss for classification and binary cross-entropy loss for reconstruction. For methods with a temperature parameter (GS and STGS), the temperature is annealed exponentially from $1$ to $0.01$.

\subsection{Compute Environment}
The experiments were run on a single Nvidia A100 GPU in a cluster environment. Mixed precision floating point operations were used wherever possible. Less than 2 GB of GPU memory is needed to train the model and we expect that our results can be reproduced on significantly less powerful hardware with reasonable training times.
\clearpage
\section{Image Reconstruction} \label{app:reconstruction}
\begin{figure*}[ht!]
    %\centering
    \scalebox{0.63}{
    \begin{subfigure}[b]{0.15\textwidth}
        \centering
        \includegraphics[width=\textwidth]{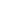}
        \includegraphics[width=\textwidth]{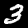}
        \includegraphics[width=\textwidth]{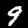}
        \includegraphics[width=\textwidth]{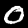}
        \caption{Ground truth}
    \end{subfigure}
    \begin{subfigure}[b]{0.15\textwidth}
        \centering
        \includegraphics[width=\textwidth]{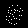}
        \includegraphics[width=\textwidth]{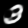}
        \includegraphics[width=\textwidth]{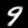}
        \includegraphics[width=\textwidth]{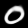}
        \caption{GS}
    \end{subfigure}
    \begin{subfigure}[b]{0.15\textwidth}
        \centering
        \includegraphics[width=\textwidth]{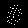}
        \includegraphics[width=\textwidth]{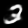}
        \includegraphics[width=\textwidth]{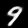}
        \includegraphics[width=\textwidth]{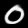}
        \caption{STGS}
    \end{subfigure}
    \begin{subfigure}[b]{0.15\textwidth}
        \centering
        \includegraphics[width=\textwidth]{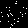}
        \includegraphics[width=\textwidth]{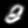}
        \includegraphics[width=\textwidth]{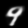}
        \includegraphics[width=\textwidth]{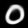}
        \caption{SFESS}
    \end{subfigure}
    \begin{subfigure}[b]{0.15\textwidth}
        \centering
        \includegraphics[width=\textwidth]{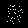}
        \includegraphics[width=\textwidth]{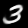}
        \includegraphics[width=\textwidth]{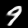}
        \includegraphics[width=\textwidth]{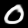}
        \caption{SFESS-V}
    \end{subfigure}}
    \setcounter{subfigure}{0}
    \hspace{4pt}
    \scalebox{0.63}{
    \begin{subfigure}[b]{0.15\textwidth}
        \centering
        \includegraphics[width=\textwidth]{figures/reconstruction/white.png}
        \includegraphics[width=\textwidth]{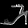}
        \includegraphics[width=\textwidth]{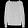}
        \includegraphics[width=\textwidth]{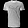}
        \caption{Ground truth}
    \end{subfigure}
    \begin{subfigure}[b]{0.15\textwidth}
        \centering
        \includegraphics[width=\textwidth]{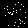}
        \includegraphics[width=\textwidth]{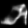}
        \includegraphics[width=\textwidth]{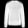}
        \includegraphics[width=\textwidth]{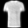}
        \caption{GS}
    \end{subfigure}
    \begin{subfigure}[b]{0.15\textwidth}
        \centering
        \includegraphics[width=\textwidth]{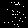}
        \includegraphics[width=\textwidth]{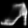}
        \includegraphics[width=\textwidth]{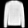}
        \includegraphics[width=\textwidth]{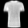}
        \caption{STGS}
    \end{subfigure}
    \begin{subfigure}[b]{0.15\textwidth}
        \centering
        \includegraphics[width=\textwidth]{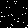}
        \includegraphics[width=\textwidth]{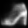}
        \includegraphics[width=\textwidth]{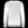}
        \includegraphics[width=\textwidth]{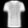}
        \caption{SFESS}
    \end{subfigure}
    \begin{subfigure}[b]{0.15\textwidth}
        \centering
        \includegraphics[width=\textwidth]{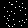}
        \includegraphics[width=\textwidth]{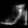}
        \includegraphics[width=\textwidth]{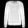}
        \includegraphics[width=\textwidth]{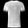}
        \caption{SFESS-V}
    \end{subfigure}}
    \setcounter{subfigure}{0}
    \scalebox{0.63}{
    \begin{subfigure}[b]{0.15\textwidth}
        \centering
        \includegraphics[width=\textwidth]{figures/reconstruction/white.png}
        \includegraphics[width=\textwidth]{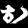}
        \includegraphics[width=\textwidth]{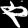}
        \includegraphics[width=\textwidth]{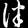}
        \caption{Ground truth}
    \end{subfigure}
    \begin{subfigure}[b]{0.15\textwidth}
        \centering
        \includegraphics[width=\textwidth]{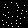}
        \includegraphics[width=\textwidth]{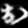}
        \includegraphics[width=\textwidth]{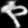}
        \includegraphics[width=\textwidth]{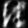}
        \caption{GS}
    \end{subfigure}
    \begin{subfigure}[b]{0.15\textwidth}
        \centering
        \includegraphics[width=\textwidth]{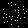}
        \includegraphics[width=\textwidth]{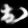}
        \includegraphics[width=\textwidth]{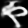}
        \includegraphics[width=\textwidth]{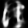}
        \caption{STGS}
    \end{subfigure}
    \begin{subfigure}[b]{0.15\textwidth}
        \centering
        \includegraphics[width=\textwidth]{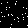}
        \includegraphics[width=\textwidth]{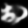}
        \includegraphics[width=\textwidth]{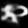}
        \includegraphics[width=\textwidth]{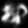}
        \caption{SFESS}
    \end{subfigure}
    \begin{subfigure}[b]{0.15\textwidth}
        \centering
        \includegraphics[width=\textwidth]{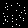}
        \includegraphics[width=\textwidth]{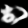}
        \includegraphics[width=\textwidth]{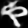}
        \includegraphics[width=\textwidth]{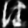}
        \caption{SFESS-V}
    \end{subfigure}}
    \caption{\textbf{Reconstruction plots}. For each dataset, the first row shows the learned selection mask and the following rows different samples from the test data. The leftmost column (a) shows the ground truth images and the following (b--e) show reconstructions from the jointly trained decoder. From top to bottom, left to right the datasets shown are MNIST, Fashion MNIST, and KMNIST.}
    \label{fig:reconstruction}
\end{figure*}

This appendix includes reconstruction plots from the MNIST, Fashion MNIST, and KMNIST test sets, which are shown in \cref{fig:reconstruction}.

\clearpage
\section{Convergence of Validation Metrics} \label{app:converge}

\begin{figure*}[ht!]
    \centering
    \includegraphics[width=0.32\textwidth]{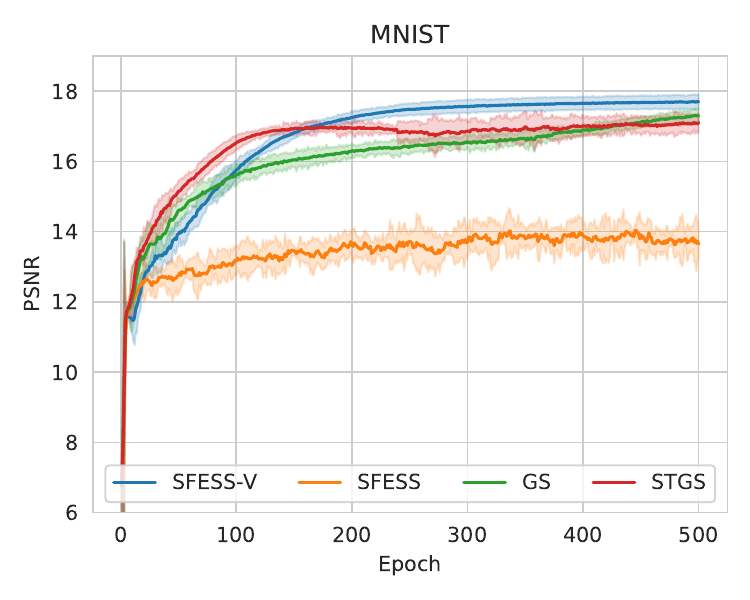}
    \includegraphics[width=0.32\textwidth]{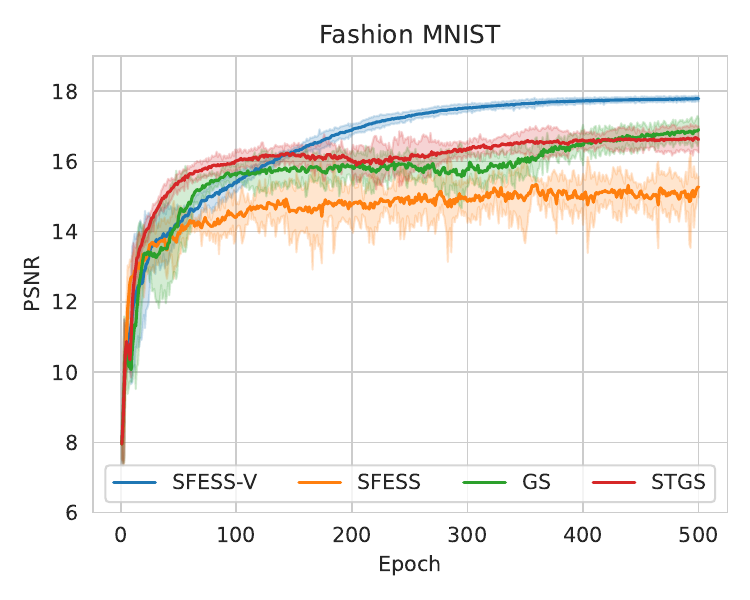}
    \includegraphics[width=0.32\textwidth]{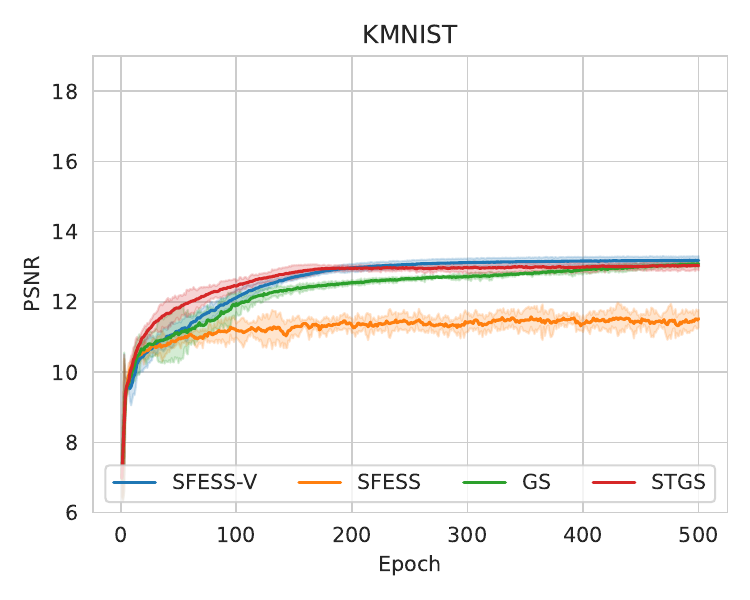}
    \includegraphics[width=0.32\textwidth]{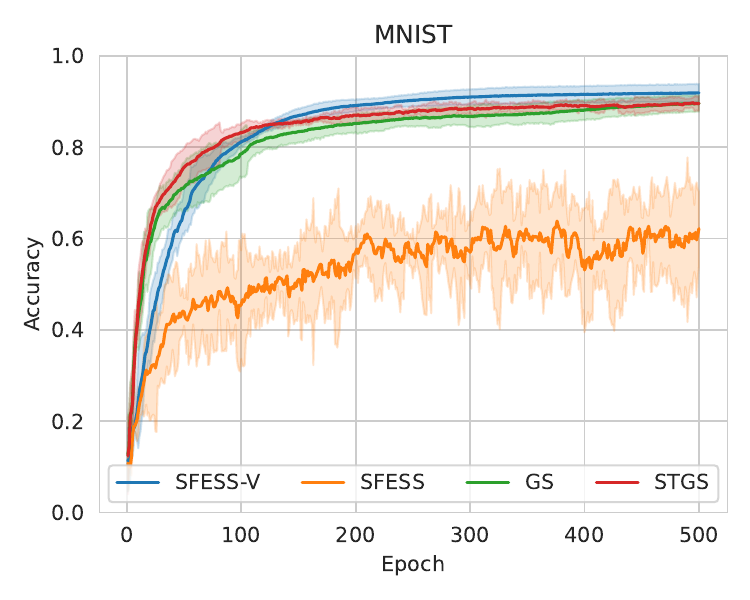}
    \includegraphics[width=0.32\textwidth]{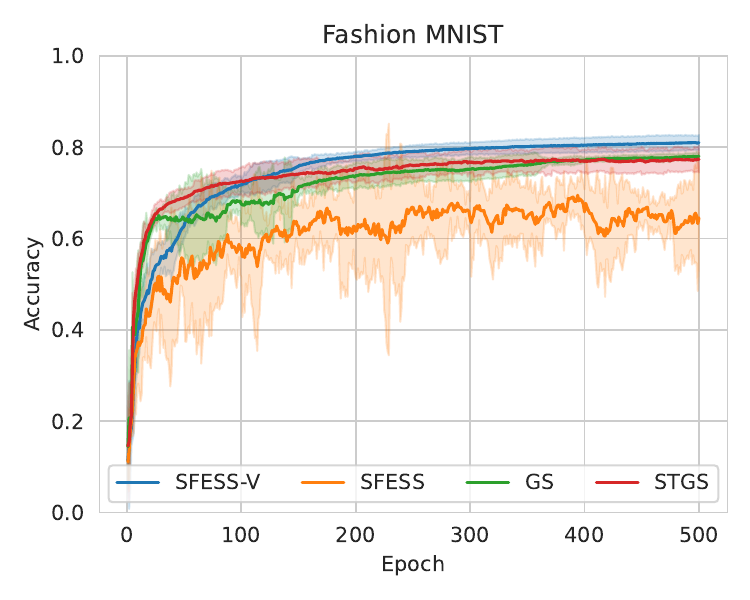}
    \includegraphics[width=0.32\textwidth]{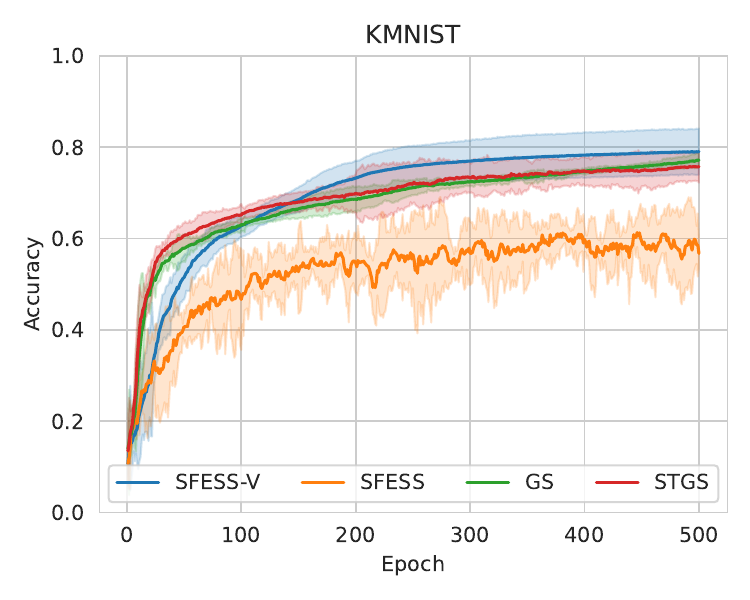}
    \caption{\textbf{Convergence plots (validation)}. Convergence of validation metrics with $k = 30$ selections. The top row shows reconstruction PSNR and the bottom row classification accuracy. The confidence intervals show two standard deviations computed for 5 repeated runs with different random seeds.}
    \label{fig:convergence-validation}
\end{figure*}
\section{Deriving the Score Function Estimator} \label{app:proof}
In this appendix, we derive the score function estimator \parencite{williams_simple_1992} which provides a Monte-Carlo estimate of the objective's gradient. We adapt the proof from \textcite{mohamed_monte_2020} (with annotations added):
\begin{align}
    \nabla_{\bm\theta} \E_{p_{\bm\theta}(\bm z)}[f_{\bm\phi}(\bm z)] &= \nabla_{\bm\theta} \sum_{\bm z} p_{\bm\theta}(\bm z)f_{\bm\phi}(\bm z) \label{eq:logderiv} &&\text{By definition of $\mathbb{E}$}\\
    &= \sum_{\bm z} \nabla_{\bm\theta} p_{\bm\theta}(\bm z)f_{\bm\phi}(\bm z) \nonumber &&\text{Interchange gradient and summation} \\ 
    &= \sum_{\bm z} p_{\bm\theta}(\bm z) \nabla_{\bm\theta} \log p_{\bm\theta}(\bm z)f_{\bm\phi}(\bm z) \nonumber &&\text{By $\log$ derivative rule}\\
    &= \E_{p_{\bm\theta}(\bm z)}[f(\bm z; \bm\phi) \nabla_{\bm\theta} \log p_{\bm\theta}(\bm z)] \label{eq:expect} &&\text{By definition of $\mathbb{E}$}\\
    &\approx \frac{1}{N} \sum_{i=1}^N f_{\bm\phi}(\bm z^{(i)})\nabla_{\bm\theta}\log p_{\bm\theta}(\bm z^{(i)}) \label{eq:mcest} &&\text{Monte-Carlo estimate}
\end{align}
By the law of large numbers, the Monte-Carlo estimator in \cref{eq:mcest} converges to the expected value in \cref{eq:expect} as $N \rightarrow \infty$, which is exactly the value of the true gradient in \cref{eq:logderiv}. Hence, the estimator is an unbiased estimator of the true gradient.
\clearpage
\section{Score Function Calculation} \label{app:score}
\begin{figure}[ht!]
    \centering
    \includegraphics[width=0.8\textwidth]{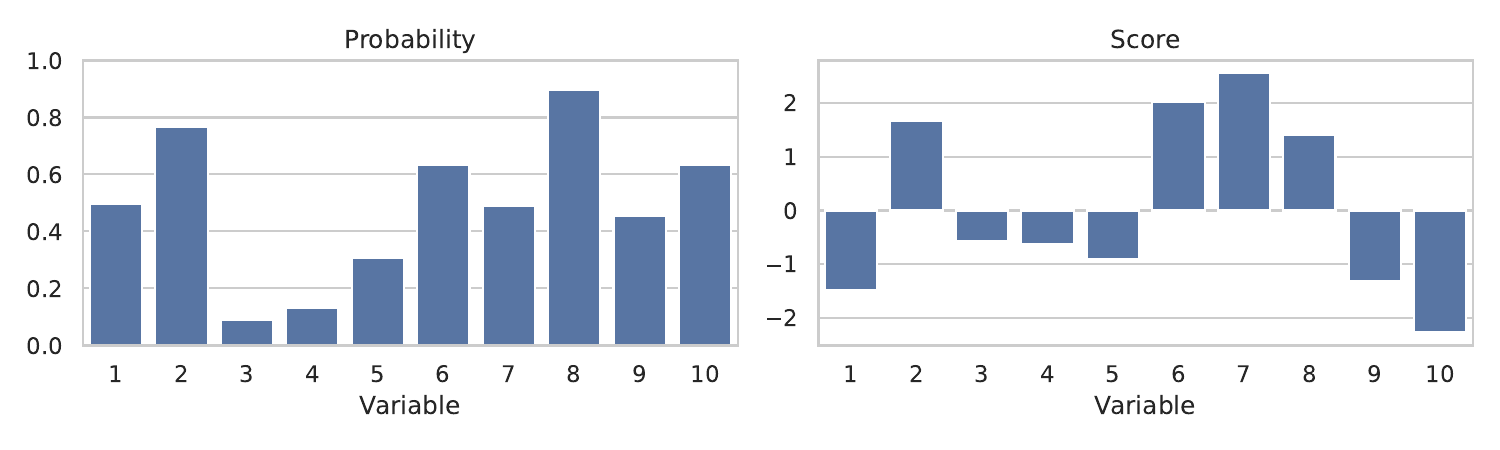}
    \caption{\textbf{Scores}. An example of scores $\nabla_{\bm\theta} \log p_{\bm\theta,k}(\bm z)$ (right) for the parameters (probabilities) $\bm\theta$ (left) for $n = 10$ and $k = 4$. It is easy to identify the sample that was evaluated -- included elements have a positive score. The scores were calculated using \cref{eq:score-function}, where the second term was computed using a DFT.}
    \label{fig:score}
\end{figure}

A key component of SFESS is calculating the score function. The unconditional independent Bernoulli distribution is renormalized by the Poisson-Binomial distribution. This renormalization factor is calculated following \textcite{fernandez_closed-form_2010}. \Cref{listing:pseudocode} outlines this calculation in pseudocode. \Cref{fig:score} shows an example of the resulting scores.

\begin{listing}[ht!]
\begin{minted}{python}
import torch
import cmath

def poibin_prob(theta, k):
    n = theta.size(0)
    i = torch.arange(n + 1).unsqueeze(-1)
    c = cmath.exp(2j * torch.pi / (n + 1))
    prod = torch.prod(theta * c**i + (1 - theta), dim=1)
    probs = torch.fft.fft(prod).real / (n + 1)
    return probs[k]
\end{minted}
\caption{PyTorch-style pseudocode for calculating the Poisson-Binomial PMF \parencite{fernandez_closed-form_2010}.}
\label{listing:pseudocode}
\end{listing}

\end{document}